\def\year{2021}\relax
\documentclass[letterpaper]{article} 
\usepackage{aaai21}  
\usepackage{times}  
\usepackage{helvet} 
\usepackage{courier}  
\usepackage[hyphens]{url}  
\usepackage{graphicx} 
\usepackage{natbib}  
\usepackage{caption} 
\title{Monitoring Diversity of AI Conferences: \\ Lessons Learnt and Future Challenges in the DivinAI Project}
 \author {
    Isabelle Hupont\textsuperscript{\rm 1},
    Emilia Gómez\textsuperscript{\rm 1},
    Songül Tolan\textsuperscript{\rm 1},
    Lorenzo Porcaro\textsuperscript{\rm 2},
    Ana Freire\textsuperscript{\rm 3} 
    \\
 }
 \affiliations {
     \textsuperscript{\rm 1} Joint Research Centre, European Commission \\
     \textsuperscript{\rm 2} Universitat Pompeu Fabra \\
     \textsuperscript{\rm 3} UPF Barcelona School of Management \\
     \{isabelle.hupont-torres, emilia.gomez-gutierrez, songul.tolan\}@ec.europa.eu \\ 
     lorenzo.porcaro@upf.edu, ana.freire@bsm.upf.edu \\
 }

\begin{document}

\maketitle

\begin{abstract}
DivinAI is an open and collaborative initiative promoted by the European Commission's Joint Research Centre to measure and monitor diversity indicators related to AI conferences, with special focus on gender balance, geographical representation, and presence of academia vs companies.
This paper summarizes the main achievements and lessons learnt during the first year of life of the DivinAI project, and proposes a set of recommendations for its further development and maintenance by the AI community.
\end{abstract}

\section{Introduction}

Diversity in a research community refers to the existence of variations of different characteristics among its members. These characteristics might be everything that makes each person unique, such as cognitive skills and personality traits, along with the factors that shape identity including race, age, gender, religion and cultural background. Evidence suggests that diverse scientific communities outperform homogeneous groups on generating new research questions, developing inclusive methodologies to better understand broader populations, problem solving, and publishing more articles with more citations~\cite{swartz2019science, alshebli2018preeminence}. Prominent Artificial Intelligence (AI) institutions and experts, such as the High Level Expert Group on AI of the European Commission~\cite{EC_ethic_guidelines} and the AI Now Institute~\cite{west2019discriminating}, have emphasized the need for fostering diversity in research and development of AI systems.

The DivinAI --Diversity in Artificial Intelligence-- project was created as a response to this need~\cite{divinAI}. DivinAI is an open and collaborative initiative promoted by the European Commission's Joint Research Centre to measure and monitor diversity indicators related to AI conferences, with special focus on gender balance, geographical representation, and presence of academia vs companies. The goal of DivinAI is addressed through a collaborative website\footnote{\url{http://www.divinai.org}} in which anyone can contribute by adding diversity data related to AI conferences worldwide. 

DivinAI was first presented in the AAAI Workshop on Diversity in Artificial Intelligence in 2021~\cite{freire2021measuring}. One year after this presentation, the platform gathers diversity indicators for a set of major AI conferences, including the AAAI Conference on Artificial Intelligence (AAAI), the International Conference on Machine Learning (ICML), the International Joint Conference on AI (IJCAI), the Conference on Neural Information Processing Systems (NeurIPS) and the Conference on Recommender Systems (RecSys). This short paper discusses the main challenges we have faced in the endeavour of measuring diversity at AI conferences throughout this year.

\section{DivinAI: Platform Overview}

DivinAI provides the AI community with an open platform to monitor diversity at AI conferences across years. It allows to compare how different AI conferences care about minorities, and to assess the impact of diversity and inclusion policies (e.g. mentoring programs, travel grants and other inclusion initiatives). This section details the diversity metrics monitored by DivinAI, as well as how data is collected and visualised.

\subsection{Diversity Metrics}

\begin{figure*}[htb!]
    \centering
    \includegraphics[width=0.9\linewidth]{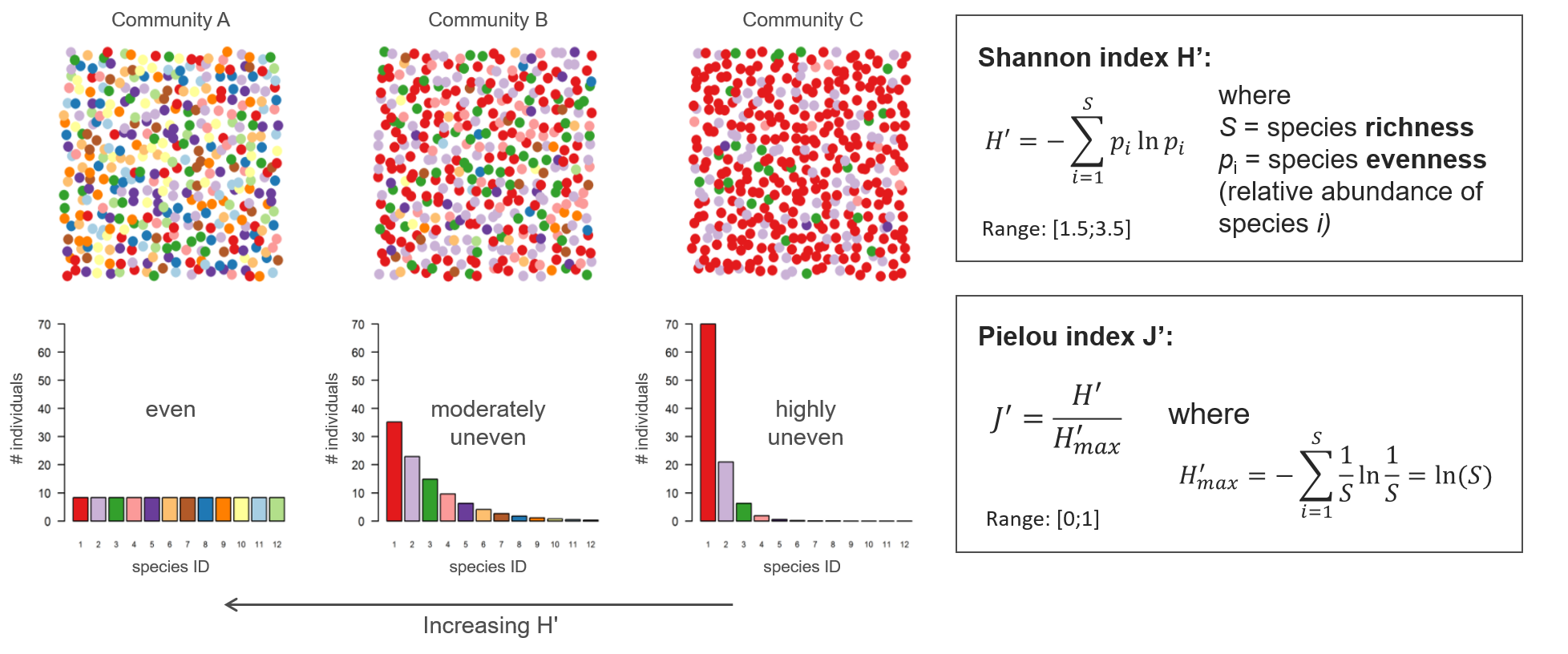}
    \caption{Formulas and illustration of the Shannon and Pielou biodiversity indexes on which DivinAI's diversity indexes are grounded.}
    \label{fig:biodiversity_idx}
\end{figure*}

DivinAI's diversity indexes are grounded on Shannon~\cite{shannon1948mathematical} and Pielou~\cite{pielou1966measurement} indexes, which are widely used in the field of plant and animal ecology to measure biodiversity. Figure~\ref{fig:biodiversity_idx} provides the formula and illustrates the meaning of these indexes. The Shannon index $H'$  increases as both the richness (number of different species) and the evenness (relative abundance of each species) of a community increases. The Pielou index $J'$ allows to compute Shannon evenness, discarding the richness factor. 

\begin{figure}[htb!]
    \centering
    \includegraphics[width=\linewidth]{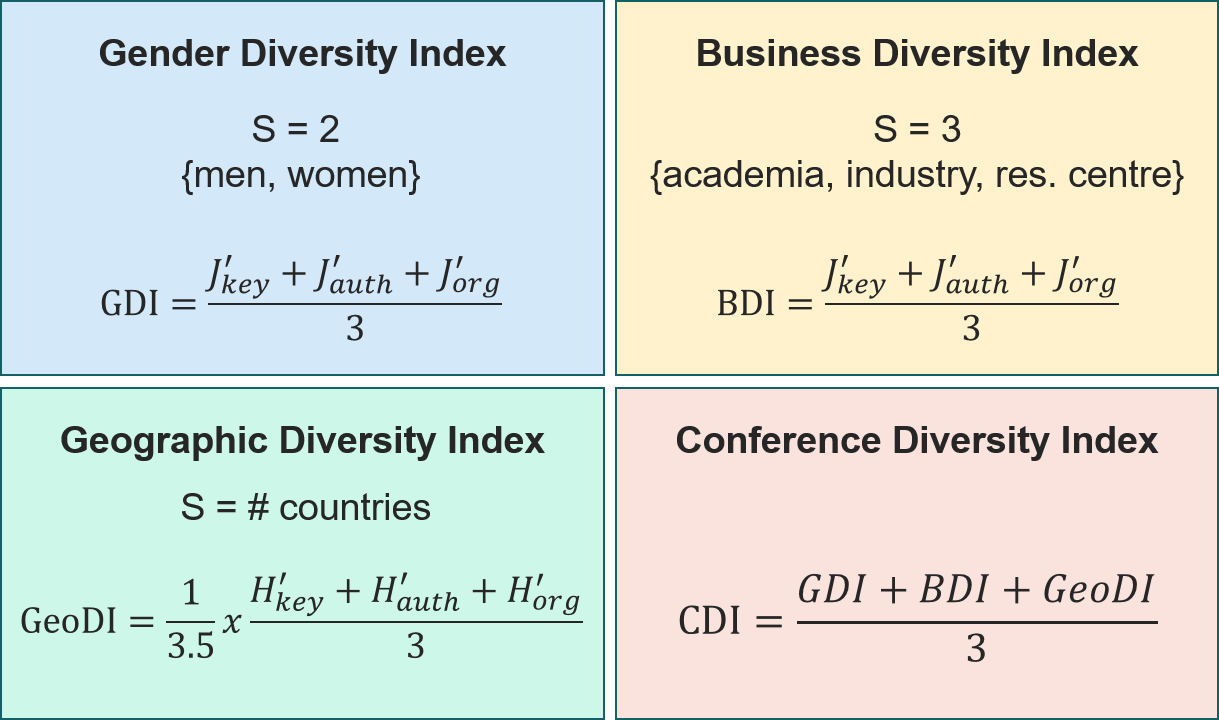}
    \caption{Formulas for the four diversity indexes monitored by DivinAI. Subindexes represent different member roles at a conference, namely: \textit{key} - keynote speakers, \textit{auth} - authors and \textit{org} - organisers.}
    \label{fig:divinai_idx}
\end{figure}

DivinAI considers three main roles that community members have within a conference: (1) keynote speakers; (2) conference organizers and technical committee members; and (3) paper authors. It also considers 3 facets of diversity: (1) gender diversity with $S=2$ groups, namely \{women,  men\}; (2) business diversity with $S=3$ groups, namely \{academia, industry, research centre\}; and (3) geographic diversity with as many groups as different countries represented by conference members. Pielou and Shannon indexes are computed for each of the three community member roles, and then averaged as indicated in Figure~\ref{fig:divinai_idx} to obtain four diversity indicators: Gender Diversity Index (GDI), Business Diversity Index (BDI), Geographic Diversity Index (GeoDI) and the overall Conference Diversity Index (CDI). For further information about how DivinAI indexes are computed, we refer the reader to~\cite{freire2021measuring,hupont2021diverse}.

\subsection{Data Collection}
\label{sec:datacollection}
The data used to compute diversity indicators for DivinAI is exclusively based on public domain information available at conference proceedings and web pages. For each paper in a conference proceedings, authors' names and individual affiliations are extracted, and gender, geographic and business information is inferred from them. The same information is collected for keynote speakers and conference organisers.

DivinAI contributors have several ways to collect conference data. For example, to foster the collaborative nature of DivinAI and raise awareness in the AI community, we organised a series of \textit{HackFests} in Barcelona and AAAI conference in New York~\cite{hackfests}. During these events, attendees manually gather and annotate data of top AI conferences for DivinAI. It is also possible to collect conference data in a semi-automated manner using the export tools offered by some publishers and bibliographic repositories such as DBLP~\cite{dblp} or Web of Science~\cite{wos}. Similarly, the gender of a given author can be manually inferred (e.g. from his/her name, or by looking at his/her profile picture online) or through automated tools such as NamSor~\cite{namsor} or Gender API~\cite{genderapi}.

\subsection{Data Visualisation}

\begin{figure*}[htb!]
    \centering
    \includegraphics[width=\linewidth]{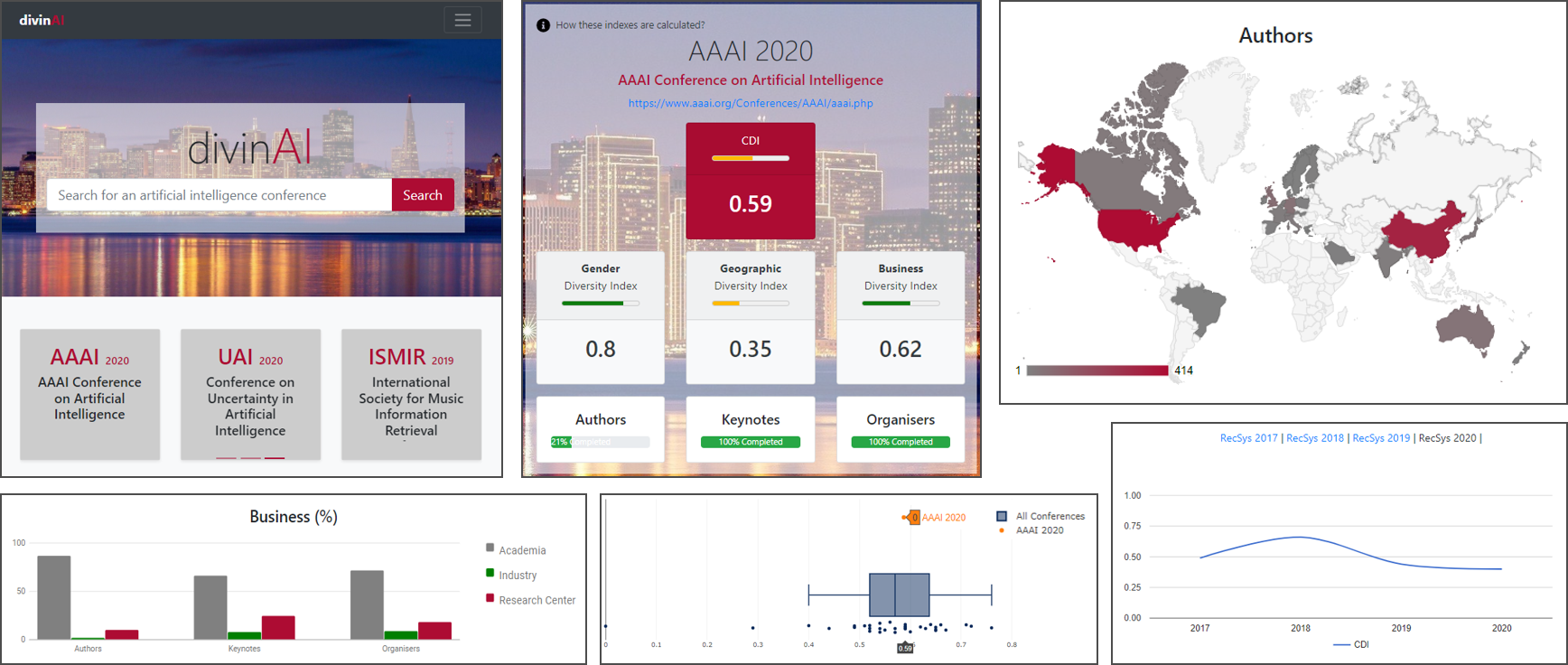}
    \caption{Main elements of DivinAI's web interface.}
    \label{fig:interface}
\end{figure*}

Figure~\ref{fig:interface} depicts the most relevant elements of DivinAI's web interface. The main menu provides a search bar to find target conferences, as well as a dynamic ruler with direct links to the different conferences available. Each conference edition (e.g. AAAI 2019, AAAI 2020, RecSys 2020, and so on) has an associated diversity dashboard showing the four diversity indexes. Additionally, other visualisations and statistics are available by scrolling down, namely: histograms with gender and business percentages; interactive world maps with the geographic distribution of authors, keynote speakers and organisers; a timeline showing the evolution of the CDI index throughout conference editions; and a boxplot positioning the current edition with regards to other AI conferences registered in the platform.

\section{Lessons learnt and recommendations}

This first year of running of DivinAI has made us learn a number of good practices and  identify limitations and paths for future work. In the following, we share with the community a set of challenges and recommendations associated to each diversity dimension. 

\subsection{Gender dimension}

As previously mentioned, gender information has been  extracted either manually (by people contributing to the project) or using automatic estimation algorithms. Being gender a very sensitive characteristic, we found several ethical challenges related to its labelling that needs to be considered. \\

\noindent\underline{Challenges:}
\begin{itemize}
\item {\bf Personal data protection vs reproducibility}. Gender is very sensitive personal data, so the association of names  and gender information should be fully protected. However, the DivinAI project wants to ensure reproducibility of the mechanisms used to compute the diversity indicators. That is, there is a trade-off between these requirements that needs to be addressed. 
\item {\bf Sex and gender: binary labeling vs continuous reality}. Names have been traditionally associated to binary (female and male) categories of sex, where sex refers to a person's biological status~\cite{Wamesley}. 
However, gender identity refers to one's internal sense of self, whether it is a man, woman, neither or both. It is important then to stress that both the manual and automatic name-based classification into binary categories is prone to errors and consequently misgendering. Being aware of the potentially harmful consequences of misgendering~\cite{dolan2020misgendering}, we aspire to do better when it comes to this task. 
\item {\bf Automatic gender labeling.} There has been some ethical concerns regarding the use of systems for the automatic labelling of gender~\cite{verge}.  Further, and in relation to the previous point, there is no study to date providing evidence on whether the quality of automatic data collection of gender is significantly different from (highly subjective) manual labeling of such data. However, the automation of the gender labeling process is essential for the task of monitoring, as we need to access data at a large scale.
\end{itemize}

\noindent\underline{Recommendations:}
\begin{itemize}
\item {\bf Strict personal data protection mechanisms} should be applied, as gender-names associations should not be shared openly. However, we may consider its storage following strict security rules for the purposes of auditing/reproducibility/accountability of the obtained results.  At the time being, all the public information in DivinAI is anonymised but we do keep records of names/surnames privately for the sake of reproducibility.
\item In order to prevent errors, gender should ideally be self-annotated. Our suggestion for this would be for conferences (similarly, bibliographic websites such as DBLP or Web of Science) to ask participants, during the registration process, if they would be willing to manually {\bf provide their own gender information}, using a rich scale, and just for diversity monitoring purposes.  
\end{itemize}

\subsection{Country dimension}

DivinAI uses authors' affiliation country --which does not necessarily match their country or even continent of origin--  as a proxy to compute the geographic diversity index (GeoDI). \\

\noindent\underline{Challenges:}

\begin{itemize}
    \item {\bf Country granularity.} DivinAI currently considers countries as the geographic level of granularity in order to capture the richness dimension in the GeoDI. As countries are highly imbalanced in terms of population, considering other granularity levels (such as continents or unions) might be equally important even though at the cost of losing the richness dimension. 
    \item \textbf{Country assignment.} Considering the affiliation country 
   instead of the origin country of authors (which is virtually impossible to infer from proceedings data) carries some political implications. The assignment by affiliation is biased against Low and Middle Income Countries (LICs/MICs) where universities and other research departments have less resources than High Income Countries (HICs) for AI research and the participation in international conferences. 
\end{itemize}

\noindent\underline{Recommendations:}

\begin{itemize}
\item The \textbf{GeoDI index should be revisited} so that it is more balanced with regard to population while keeping capturing geographic richness. 
\item As mentioned, in DivinAI we identify the country on the basis of affiliations, while remaining aware of the policy implication of this decision. A quantification of a lack of diversity, due to missing participants from LICs and MICs, points towards an inequality of financial resources. Thus, this evidence could be used to recommend a \textbf{redirection of resources towards those underrepresented countries} via research grants and scholarships. 
\end{itemize}

\vspace{4mm}

\subsection{Business dimension}

We currently assign each author to one single type of institution among research centre, industry and academia. \\

\noindent\underline{Challenges:}
\begin{itemize}
    \item {\bf Double affiliations.} We found that it is common to see authors with a double affiliation, one from an academic institution and another one from a company. At the moment we are considering the first affiliation as the most relevant one, which might not fully capture authors' actual situation.
    \item {\bf Institutions granularity}. As the AI field is becoming increasingly multi-disciplinary, more types of institutions could be covered. 
\end{itemize}

\noindent\underline{Recommendations:}
\begin{itemize}
\item Keeping in mind that the first affiliation is the one that should reflect the main institution employing the author, we should be aware of the limitations of the business diversity index to \textbf{cover multiple affiliations}. 
\item A broader spectrum of institution types could be covered by the business diversity index (BDI). In particular, we foresee to explore the \textbf{use of the Global Research Identifier Database}~\cite{GRID} to include new categories for the BDI such as healthcare centres, governments and non-profits. 
\end{itemize}

\vspace{4mm}

\subsection{General considerations}

Further, we have detected other more general challenges which are discussed below. \\ 

\vspace{12mm}

\noindent\underline{Challenges:}
\begin{itemize}
    \item {\bf Topic diversity.} We have had some requests to incorporate diversity in terms of research topics as a relevant dimension to be measured. However, research topics usually depend on the specific conference and its specialization, and there is no consensus yet in the community for the adoption of a common topic taxonomy.   
    \item {\bf Need for indicators in a meaningful temporal scope.} The experience shows the need to compute indicators for several years in a row in order to perform meaningful analyses, as different conferences may have different policies that might affect indicators such as geographical location (e.g. conferences held in Asia-Europe-USA in alternate years). 
    \item {\bf Correlation vs causality.} Similarly, although DivinAI's indicators were designed to be used as a monitoring tool of diversity initiatives, they might be highly influenced by other factors than diversity policies, such as the conference location (virtual or physical, highly related to geographical diversity indicators). 
    \item {\bf Meaningfulness of indicators for communication purposes.} Are simple percentages easier to understand than our proposed diversity indicators? 
    \item {\bf Incentives for conferences to contribute.} Ideally, conference organisers should make it customary to collect and upload diversity data to the platform, which has unfortunately not yet happened to date (the DivinAI team has mostly been the one in charge of uploading the data based on what is published in public proceedings). The reason is probably the lack of incentive to do so, especially as data collection is a tedious process. The active contribution of conference representatives would allow for a more comprehensive analysis and to consider other attendees (i.e. non-authors, -keynotes or -organisers) that represent a critical point of view on the diversity of a conference.
\end{itemize}

\noindent\underline{Recommendations:}
\begin{itemize}
\item We plan to incorporate a \textbf{topic diversity index} to DivinAI in the near future. For that purpose, we will explore different methodological approaches inspired by current work on AI taxonomies, where AI taxonomies are defined based on a review of existing call for papers from conferences and keywords used in different research platforms~\cite{Samoili2021}. 
\item \textbf{Longitudinal analyses} are needed to fully assess the real impact of --still very recent-- diversity initiatives. Monitoring through DivinAI should be maintained for at least the next decade to start getting meaningful insights.
\item Further assessment is needed on proper \textbf{communication strategies} to disseminate DivinAI's results to reach not only AI practitioners but also other stakeholders, such as the wide public and policy makers. 
\item We should research on potential \textbf{incentivation mechanisms} for conferences to contribute, so that we can ensure the data continuity, completeness and project sustainability.
\end{itemize}

\section{Conclusions}

This paper summarizes the main achievements and lessons learnt during the first year of life of the DivinAI project. While the platform has been successfully used as a tool to collect historical diversity data from the largest conferences on Artificial Intelligence (AAAI, ICML, IJCAI, NeurIPS, RecSys, among others), its long-term continuity is needed to monitor the real impact of diversity policies --that have just begun to shed the light-- and identified challenges must be carefully addressed.
In the near future, we hope to implement the proposed recommendations and gather additional feedback from the community linked to the workshop to further promote the adoption of the platform at a large and longitudinal scale.

\bibliography{references.bib}

\section{ Acknowledgments}
This work is partially supported by the European Commission under the HUMAINT project of the Joint Research Centre. 

\end{document}